\pdfoutput=1

\documentclass[11pt]{article}

\usepackage[preprint]{acl}

\usepackage{times}
\usepackage{latexsym}

\usepackage[T1]{fontenc}

\usepackage[utf8]{inputenc}

\usepackage{microtype}

\usepackage{inconsolata}

\usepackage{microtype}
\usepackage{graphicx}
\usepackage{booktabs}
\usepackage{float}
\usepackage{subfig}

\usepackage{tikz}
\usepackage{pgfplots}
\usepackage{pgfplotstable}

\usepackage{amsmath}
\usepackage{amssymb}
\usepackage{mathtools}
\usepackage{amsthm}

\usepackage[capitalize,noabbrev]{cleveref}
\theoremstyle{plain}

\theoremstyle{definition}

\theoremstyle{remark}

\newcommand{\minisection}[1]{\noindent{\textbf{#1}}.}
\usepackage{multirow}
\usepackage{tabularx}
\usepackage{bbm}
\usepackage{makecell}
\usepackage{enumitem}

\usepackage{moresize}

\usepackage{threeparttable}

\usepackage{pifont}
\usepackage{fontawesome}

\title{Divergent Thoughts toward One Goal: LLM-based Multi-Agent Collaboration System for Electronic Design Automation}

\author{
\textbf{Haoyuan Wu\textsuperscript{$\spadesuit$}},
\textbf{Haisheng Zheng\textsuperscript{$\heartsuit$}},
\textbf{Zhuolun He\textsuperscript{$\spadesuit$,$\clubsuit$}},
\textbf{Bei Yu\textsuperscript{$\spadesuit$}}
\\\\
\textsuperscript{$\spadesuit$}The Chinese University of Hong Kong, Hong Kong SAR \\
\textsuperscript{$\heartsuit$}Shanghai Artificial Intelligent Laboratory, China \\
\textsuperscript{$\clubsuit$}ChatEDA Tech, China\\
\texttt{\{hywu24,byu\}@cse.cuhk.edu.hk} \\
}

\begin{document}
\maketitle

\begin{abstract}
Recently, with the development of tool-calling capabilities in large language models (LLMs), these models have demonstrated significant potential for automating electronic design automation (EDA) flows by interacting with EDA tool APIs via EDA scripts.
However, considering the limited understanding of EDA tools, LLMs face challenges in practical scenarios where diverse interfaces of EDA tools exist across different platforms.
Additionally, EDA flow automation often involves intricate, long-chain tool-calling processes, increasing the likelihood of errors in intermediate steps.
Any errors will lead to the instability and failure of EDA flow automation.
To address these challenges, we introduce EDAid, a multi-agent collaboration system where multiple agents harboring divergent thoughts converge towards a common goal, ensuring reliable and successful EDA flow automation. 
Specifically, each agent is controlled by ChipLlama models, which are expert LLMs fine-tuned for EDA flow automation.
Our experiments demonstrate the state-of-the-art (SOTA) performance of our ChipLlama models and validate the effectiveness of our EDAid in the automation of complex EDA flows, showcasing superior performance compared to single-agent systems.
\end{abstract}

\section{Introduction}
Electronic design automation (EDA) is indispensable for the design of integrated circuits (ICs).
EDA tools are integrated into a complex design flow and utilize programming interfaces to control the design process.
EDA platforms such as OpenROAD~\cite{ajayi2019openroad} and iEDA~\cite{li2024ieda}, consist of complex procedures with various configurations.
Circuit design engineers utilize EDA tools iteratively to fulfill design targets, relying on tailored scripts that manipulate these tools via programming interfaces.
However, interacting with EDA tools through scripting~\cite{chen2001scripteda} is often laborious and error-prone. 
This complexity is further intensified when design teams employ tools from various vendors in the circuit design process.

Large language models (LLMs)~\cite{openai2023gpt4, 2024claude, dubey2024llama3} have demonstrated profound instruction comprehension, planning, and reasoning capabilities. 
The potential of LLMs to interact with diverse tools for executing complex tasks has gained increasing recognition~\cite{qin2023toolllm}. 
Researchers have explored the automation of complex EDA flows by interfacing with EDA tools via LLMs~\cite{wu2024chateda, liu2023chipnemo}.
Specifically, ChatEDA~\cite{wu2024chateda} uses LLMs as ``brains'' of the agent to generate EDA scripts, automating EDA tool utilization, reducing the workload of circuit design engineers, and minimizing errors.

Although LLMs have shown potential in EDA flow automation, significant challenges remain.
Firstly, although LLMs excel at understanding natural language, they lack specialized knowledge of EDA tool usage.
These tools are designed for specific tasks such as logic synthesis, floorplanning, placement, and routing, each requiring detailed domain-specific knowledge and familiarity with various EDA tool interfaces.
To solve these problems, LLMs can be fine-tuned on datasets containing tutorials and EDA scripts specific to EDA tools~\cite{wu2024chateda}.
However, this approach presents its problems.
Each EDA platform has a unique set of commands and flows that must be mastered for effective use. 
If LLMs only focus on a specific EDA tool or platform during the instruction tuning process will limit their cross-platform utility, potentially reducing the effectiveness in practical scenarios.
Furthermore, EDA flows typically involve a sequence of intermediate steps. 
A single error in any of these steps can lead to failure in the overall process.
This risk is compounded by the probabilistic nature of LLMs, which may produce varying solutions to the same task. 
Moreover, errors may occur in intermediate steps during a long-chain tool-calling process, introducing instability into the EDA flow automation.

To address these challenges, we introduce a multi-agent collaboration system, EDAid, designed for EDA flow automation through script generation in response to natural language instructions. 
This system is characterized by the collaboration of multiple agents, which are powered by the LLMs.
As mentioned earlier, EDA flow automation is challenging even for the greatest LLM such as GPT-4~\cite{openai2023gpt4}. 
Consequently, we develop ChipLlama models, expert LLMs fine-tuned for EDA flow automation. 
We specifically focus on improving the understanding of overall EDA flow rather than simple EDA tool usage during the fine-tuning process.
We also employ few-shot chain-of-thought (CoT) prompts for each agent to further develop the performance and portability through the in-context learning capability of LLMs. 
Based on multiple agents, we propose the multi-agent system, EDAid, to ensure stability and avoid erroneous intermediate steps during the long-chain EDA tool calling process.
In this system, multiple agents collaborate with divergent thoughts following different few-shot contexts and then make the final decision based on divergent thoughts, working in concert to automate the EDA process.
Our EDAid can interpret human instructions, plan EDA tasks, and interface with EDA tools through APIs, serving as a valuable assistant in automating EDA flows and eliminating the need for manual intervention.
In summary, our contributions are as follows:
\begin{itemize}[itemsep=0pt,topsep=0pt,parsep=0pt]
	\item We develop the ChipLlama-powered agent, collaborating with few-shot CoT prompts, to develop the performance and portability of the single-agent system;
    \item Propose EDAid, a multi-agent system that collaborates multiple agents including divergent-thoughts agents and a decision-making agent for EDA flow automation;
    \item Perform extensive evaluations, which demonstrate the SOTA performance of ChipLlama models, the effectiveness of the few-shot CoT prompting, and the superior performance of our EDAid for EDA flow automation. 
\end{itemize}

\section{ChipLlama-powered Agent}

\begin{figure}[tb!]
    \centering
    \includegraphics[width=0.96\linewidth]{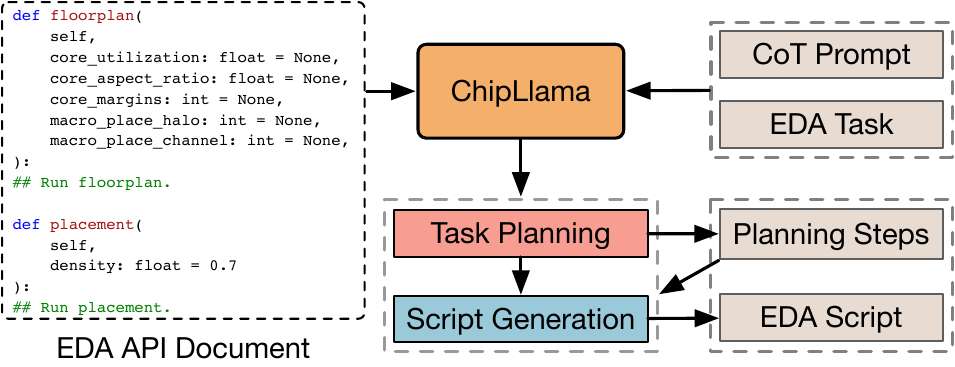} 
    \caption{Overview of the ChipLlama-powered agent for task planning and EDA script generation. 
    } 
    \label{fig:ChipAgent}
\end{figure}

For an autonomous agent for EDA flow automation, users can provide the EDA task in natural language, and the agent will generate executable scripts to complete the EDA task via an LLM.
To guarantee the performance and reliability of the overall flow, we introduce the expert LLM, ChipLlama, as the ``brain'' of the single-agent system.  
Furthermore, we apply few-shot CoT prompts to the single-agent system to enhance the reasoning ability for EDA flow automation as shown in \Cref{fig:ChipAgent}.

\subsection{ChipLlama for EDA Flow Automation}

ChipLlama models are expert LLMs fine-tuned based on Llama3~\cite{dubey2024llama3} via the hybrid instruction tuning.
The capability of the ChipLlama models determines the performance of EDA task-resolving.
Previous expert LLMs for EDA flow automation (e.g. AutoMage2~\cite{wu2024chateda}) only show reliable performance on a single platform.
However, there are various EDA platforms in real industrial scenarios.
To address this challenge, we enhance the generalization ability of LLMs by broadening their understanding of the entire EDA flow, beyond mere simple tool usage.

The invocation of EDA tools is dependent on maintaining the correct logical order in contrast to other fields where tool usage may not require a strict logical sequence. 
This dependency arises from the nature of EDA processes, which consist of interconnected stages such as logic synthesis, floorplanning, placement, and routing. 
Each stage must be executed sequentially, relying on precise inputs from preceding steps.
Therefore, besides domain-specific knowledge for EDA tool usage, LLMs must exhibit advanced logical reasoning and tool manipulation (via code) skills.

\begin{figure}[tb!]
    \centering
    \includegraphics[width=0.96\linewidth]{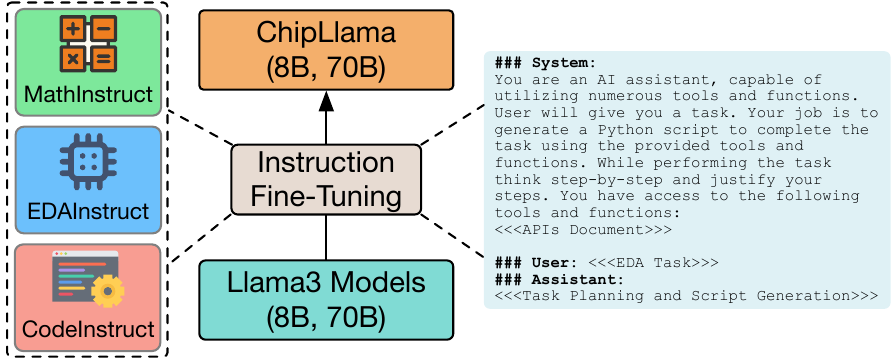} 
    \caption{Overview of hybrid instruction tuning.}
    \label{fig:hit}
\end{figure}

Instruction tuning is founded on the principle that by engaging in supervised learning driven by task-specific instructions, LLMs can acquire the skill to adhere to directives for tasks they have not previously encountered. 
This facilitates the application of LLMs to EDA tasks utilizing datasets from domains beyond EDA. 
Instruction datasets from various fields provide a wealth of directives, promoting the ability of models to develop versatile problem-solving strategies by correlating inputs with outputs.
Consequently, as illustrated in \Cref{fig:hit}, we utilize hybrid instruction tuning for ChipLlama models, which integrates three specially curated datasets including MathInstruct~\cite{yu2023metamath}, CodeInstruct~\cite{wei2023magicoder}, and EDAInstruct~\cite{wu2024chateda}.  
Hybrid instruction tuning fosters LLMs' deep understanding of sophisticated EDA flow automation, which can be generalized to various EDA platforms and enhance the performance and reliability applying to a single EDA platform.
We show more details of hybrid instruction tuning in \Cref{sec:appendix1}.

\subsection{Few-shot CoT Prompting}
\label{sec:ChipAgent-CoT}

Backend design involves complex procedures to interact with various EDA tools.
For example, achieving timing closure has to optimize cell placement, clock trees, and signal routing iteratively until satisfying performance is obtained.
In this sense, it is difficult for LLMs to generate a script that finishes the job directly~\cite{wu2024chateda}.
The enhancement for LLMs in such script generation can be achieved through the application of CoT prompting~\cite{wei2022cot}.
As illustrated in \Cref{fig:ChipAgent}, after receiving user instruction and API document, ChipLlama models start to plan how to complete the task in several steps following the CoT prompt and then generate the corresponding script according to the planning steps. 

Let's first focus on the standard prompt for EDA task-solving without the task planning phase. 
For standard prompting, the objective is to maximize the probability of the script $\mathcal{A}$ given an EDA task $\mathcal{Q}$, a prompt $\mathcal{T}$, and the probabilistic LLM $p_{\mathcal{L}}$ representing the utilized ChipLlama model. The probability is expressed as:
\begin{equation}
    p(\mathcal{A}|\mathcal{Q},\mathcal{T}) = \prod \limits_{i=0}^{|\mathcal{A}|} p_{\mathcal{L}}(a_{i}|\mathcal{Q},\mathcal{T},\mathcal{A}_{<i}),
    \label{Eq:standard_prompt}
\end{equation}
where $\mathcal{A}_{<i} = \{a_1, a_2, \cdots a_{i-1}\}$, $a_{i}$ represents the $i$-th token and $|\mathcal{A}|$ denotes the length of the EDA script $\mathcal{A}$. 

\begin{figure}[tb!]
    \centering
    \includegraphics[width=0.96\linewidth]{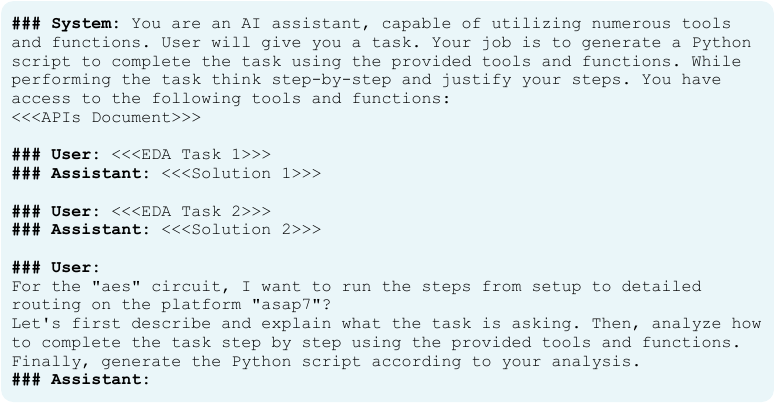} 
    \caption{Few-shot CoT prompt template utilized in ChipLlama-powered agent.}
    \label{fig:cot_prompt}
\end{figure}

\begin{figure*}[tb!]
    \centering
    \includegraphics[width=0.98\linewidth]{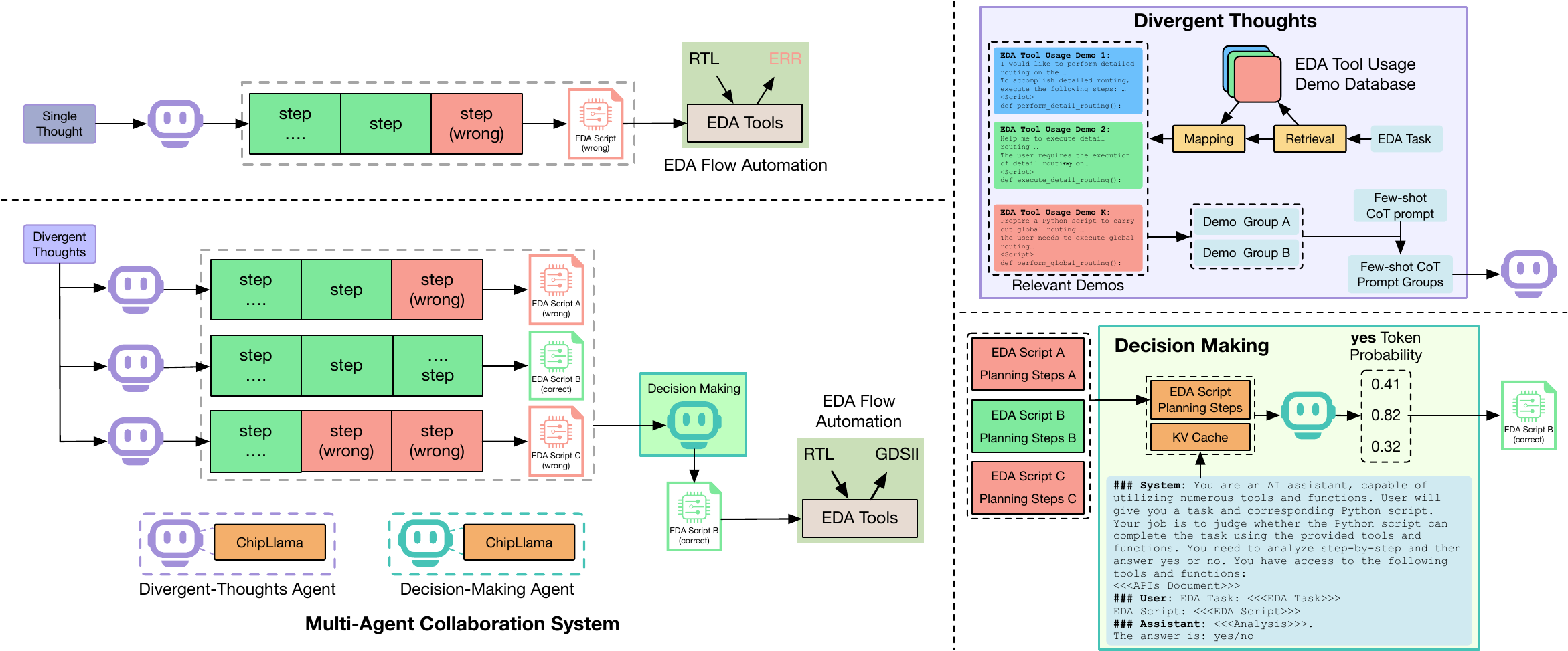} 
    \caption{Overview of EDAid, the multi-agent collaboration system. Given an EDA task, multiple agents (including divergent-thoughts agents (role $R_{0}$) and a decision-making agent (role $R_{1}$)) collaborate to generate the EDA script. Finally, the generated EDA script will automate the EDA flow interfacing the EDA tools via APIs.}
    \label{fig:multi_agent_cosys}
\end{figure*}

Zero-shot CoT prompting is a straightforward concept that incorporates a sequence of task planning steps into the initial prompt.
In this scenario, $p_{L_{0}}$ processes $\mathcal{Q}$ and $\mathcal{T}$ to generate a models of planning steps $\mathcal{C}$ and then generate the EDA script $\mathcal{A}$. 
In addressing $\mathcal{Q}$, articulated in natural language, the expert LLM from the ChipLlama models dissects it into a sequential set of steps using $\mathcal{V}$ as per the guidelines outlined in the API document within $\mathcal{T}$. 
This decomposition facilitates streamlined handling through the utilization of EDA tools. 
Furthermore, meticulous determination of parameters required at each step occurs during the task planning phase, minimizing the potential for erroneous parameter usage in EDA script generation. 
Following the task planning phase, structured steps $\mathcal{C}$ are formulated, enhancing the efficient orchestration of the intricate EDA task. 
Each step is executable through the corresponding APIs of the EDA tools.
Subsequently, LLMs can formulate the script $\mathcal{A}$ to invoke these APIs for automating the EDA flow.
Therefore, \Cref{Eq:standard_prompt} can be modified to: 
\begin{equation}
    p(\mathcal{A}\mid\mathcal{Q},\mathcal{T}) = p(\mathcal{A}\mid\mathcal{Q},\mathcal{T},\mathcal{C})p(\mathcal{C}\mid\mathcal{Q},\mathcal{T}),
    \label{Eq:zero-shot_CoT}
\end{equation}
where $p(\mathcal{A}\mid\mathcal{Q},\mathcal{T},\mathcal{C})$ and  $p(\mathcal{C}\mid\mathcal{Q},\mathcal{T})$ are defined as follows:
\begin{align}
    p(\mathcal{C}\mid\mathcal{Q},\mathcal{T}) &= \prod \limits_{i=0}^{|\mathcal{C}|} p_{\mathcal{L}}(c_{i}\mid\mathcal{Q},\mathcal{T},\mathcal{C}_{<i}),
    \label{Eq:zero-shot_CoT_step1} \\
    p(\mathcal{A}\mid\mathcal{Q},\mathcal{T},\mathcal{C}) &= \prod \limits_{i=0}^{|\mathcal{A}|} p_{\mathcal{L}}(a_{i}\mid\mathcal{Q},\mathcal{T},\mathcal{C},\mathcal{A}_{<i}).
    \label{Eq:zero-shot_CoT_step2}
\end{align}
Here, $\mathcal{C}_{<i} = \{c_1, c_2, \cdots c_{i-1}\}$.
$c_{i}$ and $|\mathcal{C}|$ indicate the $i$-th token and the length of the task planning steps, respectively. 
Zero-shot CoT prompt is provided to guide LLMs in generating task planning steps $\mathcal{C}$ before generating the EDA script $\mathcal{A}$.

Few-shot CoT prompting merges the paradigms of in-context learning with zero-shot CoT prompting to enhance performance on complex EDA tasks that necessitate planning guidelines for resolution. 
In few-shot CoT scenario, $\mathcal{T} = {(\mathcal{Q}_{i}, \mathcal{C}_{i}, \mathcal{A}_{i})}^{\mathcal{N}}_{i=1}$, which consists of $\mathcal{N}$ instances of $(\mathcal{Q}, \mathcal{C}, \mathcal{A})$ tuple. 
For an instance $(\mathcal{Q}, \mathcal{C}, \mathcal{A})$, this instance serves as a guide for resolving the EDA task.
Initially, it directs LLMs to decompose the EDA task $\mathcal{Q}$ according to the API document.
Following the decomposition steps, encompassing logic synthesis, floorplan creation, placement, clock tree synthesis (CTS), routing, and other relevant processes, is generated within the context of $\mathcal{C}$.
Ultimately, the instance guides LLMs on generating the EDA script $\mathcal{A}$ based on the decomposed steps $\mathcal{C}$.
As shown in \Cref{fig:cot_prompt}, a few-shot CoT prompt is provided to guide LLMs on how to generate task planning steps $\mathcal{C}$ and the EDA script $\mathcal{A}$ following previous EDA tool usage demos.

LLMs can acquire the skills to plan for an EDA task and generate the corresponding script by learning from a set of $\mathcal{N}$ instances in a few-shot CoT prompt $\mathcal{T}$. 
Moreover, consider the scenario where a design team has acquired a brand new EDA tool.
By providing ChipLlama models with brand new EDA APIs, they can still draw knowledge for EDA flow automation from few-shot CoT prompts.

\section{Multi-Agent Collaboration System}
\label{sec:multi-ChipAgent}

Automating EDA flows within complex real-world scenarios poses a formidable challenge for single-agent systems (shown in \Cref{fig:cases1}). 
These complexities arise from the multidimensional nature of EDA projects, which demand technical expertise and the need for a long-chain EDA tool-calling process.
As shown in \Cref{fig:multi_agent_cosys}, the single-agent system will make mistakes in the intermediate steps during the complex EDA task planning process, leading to failure in EDA flow automation.
In response, we introduce EDAid, a collaborative multi-agent system designed for EDA flow automation. 
Unlike the single-agent system utilizing the vanilla CoT, this system features multiple agents operating in synergy with self-consistency to resolve more complex EDA tasks.
In the following subsections, we will elucidate the mechanisms within the multi-agent system, which coordinate the efforts of multiple LLM-powered agents to automate the EDA flow.

\subsection{Multi-Agent System Definition}
For clear clarification, we conceptualize the operational environment of the multi-agent system as a graph $\mathcal{G}=(\mathcal{V}, \mathcal{E})$, where $\mathcal{V}$ represents the nodes corresponding to the agents, and $\mathcal{E}$ signifies the edges that delineate the communicative links among the agents.

\minisection{Agent Definition} 
Each agent $i \in \mathcal{V}$ is defined by a tuple $\mathcal{V}_{i} = (L_{i}, R_{i})$. 
Here, $L_{i}$ denotes the specific instances of LLMs employed by the agent, including their types and configurations, and the different prompt designs for different LLMs. 
$R_{i}$ designates the agent's role, which determines its duties and imparts a defined purpose and direction, thereby steering its actions and interactions. 
As shown in \Cref{fig:multi_agent_cosys}, two distinct roles are identified.
Specifically, divergent-thoughts agents (role $R_{0}$) are tasked with comprehending user inquiries and generating scripts for EDA flow automation.
The decision-making agent (role $R_{1}$) is the decision-maker who selects the optimal solution from various solutions to resolve the EDA task.

\minisection{Connection and Message Definition} 
Each edge $e_{ij} \in \mathcal{E}$ establishes a communicative link between agent $\mathcal{V}_{i}$ and $\mathcal{V}_{j}$, facilitating message exchange. 
A message $m$ encapsulates the content that includes the task plan and the relative EDA script, which can be transmitted from agent $\mathcal{V}_{i}$ to $\mathcal{V}_{j}$ via the established channel $e_{ij}$. 


\subsection{Divergent Thoughts}
Considering the superior performance of in-context learning, most tasks in various domains can be resolved by LLMs in one go. 
However, LLMs exhibit errors in EDA flow automation considering their multiple complex planning steps.
A notable characteristic of human cognition is the diversity of thought processes. 
Similarly, multiple planning pathways can be employed to address EDA tasks. 
Single-agent systems are not infallible in task planning, especially for scenarios that require long-chain tool-calling processes.
As depicted in \Cref{fig:multi_agent_cosys}, a single-agent system may occasionally pursue incorrect planning steps or commit errors within the task planning process. 
Specifically, a single-agent system might engage in placement without preceding floorplanning execution, or it could employ parameters tailored for the CTS stage at the placement phase.
Such flawed planning can not converge on the correct EDA script (shown in \Cref{fig:cases1}). 
Consequently, multiple planning pathways searching via multiple divergent-thoughts agents is necessary.

According to \Cref{Eq:zero-shot_CoT_step1}, planning steps $\mathcal{C}$ are contingent upon the prompt $\mathcal{T}$ when employing greedy decoding in CoT prompting. 
By supplying prompts derived from distinct distributions, we can engender a variety of planning steps $\mathcal{C}$, which facilitates the generation of answers $\mathcal{A}$ from divergent-thoughts agents that reflect divergent thinking.
In the context of few-shot CoT prompts, where $\mathcal{T} = {(\mathcal{Q}_{i}, \mathcal{C}_{i}, \mathcal{A}_{i})}^{\mathcal{N}}_{i=1}$, employing different permutations of $(\mathcal{Q}, \mathcal{C}, \mathcal{A})$ tuples can lead to the generation of $\mathcal{O} = \{\mathcal{O}_{1}, ..., \mathcal{O}_{i}\}$ that embody divergent thoughts.
The divergent thoughts process consists of two core components: relevant EDA tool usage demos retrieval and generation of different few-shot prompts with varied $(\mathcal{Q}, \mathcal{C}, \mathcal{A})$ tuples.

\minisection{Relevant EDA Tool Usage Demos Retrieval}
As shown in \Cref{fig:multi_agent_cosys}, the EDA tool usage demo database contains numerous instances, each of which includes a $(\mathcal{Q}, \mathcal{C}, \mathcal{A})$ tuple. 
The embedding model encodes each EDA task $\mathcal{Q}$ into a vector, creating an EDA task vector database.
For a new EDA task, the cosine similarity is computed between its embedding and all vectors in the vector database.
The Top-$K$ most similar tasks are identified, and their corresponding IDs are mapped back to the demo database to retrieve the $K$ most relevant instances with their respective $(\mathcal{Q}, \mathcal{C}, \mathcal{A})$ tuples.

\minisection{Different Few-shot Prompts Generation}
As shown in \Cref{fig:multi_agent_cosys}, several instances are randomly chosen from the retrieved pool of $K$ relevant demos to create a demo group. 
This random selection process is iterated to produce multiple demo groups, such as Group A, Group B, Group C, etc.
Each demo group is concatenated with the few-shot prompt template (displayed in \Cref{fig:cot_prompt}), yielding several few-shot prompt groups with different prompts $\mathcal{T} = {(\mathcal{Q}_{i}, \mathcal{C}_{i}, \mathcal{A}_{i})}^{\mathcal{N}}_{i=1}$.
Upon inputting these differing few-shot prompts into divergent-thoughts agents, they generate various outputs $\mathcal{O}$ according to the given prompts, thus realizing the objective of divergent thoughts.

\subsection{Decision Making}
\label{sec:multi-ChipAgent-dm}

Multiple divergent-thoughts agents $\mathcal{V}_{i}$ generate diverse outcomes $\mathcal{O}_{i} = (\mathcal{C}_{i}, \mathcal{A}_{i})$, reflecting divergent thoughts with different few-shot prompts. 
These outcomes $\mathcal{O}_{i} \in \mathcal{O}$ are encapsulated within a message $m = \{\mathcal{O}_{1}, ..., \mathcal{O}_{i}\}$ and subsequently dispatched to the decision making agent $\mathcal{V}_{j}$ with role $R_{1}$. 
Upon receipt, $\mathcal{V}_{j}$ start to perform a multiple choice selection with various $\mathcal{O}_{i}$ as choices, and determines which $\mathcal{O}_{i}$ should be selected as the representative output of the multi-agent system. 
As depicted in \Cref{fig:multi_agent_cosys}, the generated EDA script will automate the EDA flow interfacing with EDA tools.

As illustrated in \Cref{fig:multi_agent_cosys}, we combine the EDA task and prompt into a complete input and feed it into the decision-making agent for multiple choice selection. 
Specifically, we ask the decision-making agent to analyze and judge whether the generated EDA scripts can solve the EDA task.
Next, we calculate the probability of selecting the \textbf{yes} token from $\{\text{\textbf{yes}}, \text{\textbf{no}}\}$ at the end of the answer for each candidate EDA script as shown in \Cref{fig:multi_agent_cosys}.
Finally, we select the candidate answer with the highest \textbf{yes} token probability as the representative output of the EDAid.
Moreover, considering that the input part is the same for different candidate answers, we store the system prompt in the KV Cache to avoid redundant computation.	
\section{Experiments}
\label{sec:exp}

\subsection{Experiments Setting}
We utilize a comprehensive evaluation benchmark ChatEDA-bench~\cite{wu2024chateda}, which comprises 50 distinct tasks with target APIs from OpenROAD~\cite{ajayi2019openroad}, to evaluate the performance of our EDAid and ChipLlama models.
Moreover, we also design a benchmark, iEDA-bench, based on iEDA~\cite{li2024ieda} comprising 50 distinct tasks to evaluate the generalization to any EDA tools on different platforms of ChipLlama.
ChatEDA-bench and iEDA-bench use the accuracy of the generated EDA script as the evaluation metric.
Specifically, accuracy is related to the successful EDA flow automated through the correct generated EDA script.

\begin{figure}[!tb]
    \centering
    \includegraphics[width=0.98\linewidth]{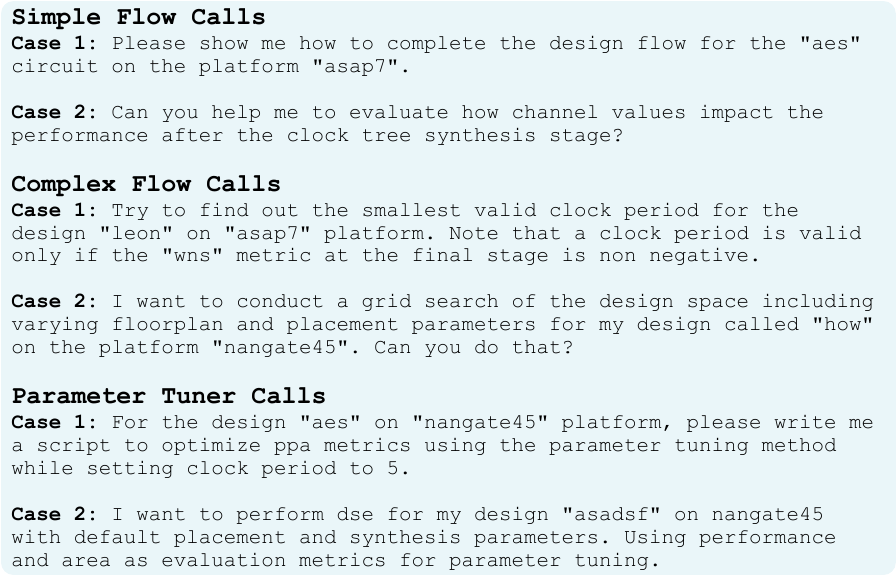} 
    \caption{Examples of evaluation benchmarks. 
    } 
    \label{fig:exp_bench}
\end{figure}

\subsection{Examples of Evaluation Benchmarks}
\label{sec:eval}
As illustrated in \Cref{fig:exp_bench}, we show some examples of our evaluation benchmarks including ChatEDA-bench~\cite{wu2024chateda} and iEDA-bench, both of which are comprehensive evaluation benchmarks comprising 50 distinct tasks including three distinct categories: simple flow calls (30\%), complex flow calls (30\%), and parameter flow calls (40\%). 

\minisection{Simple Flow Calls}
This task requires the successful execution of the whole process, including evaluation.
These cases test the fundamental application of LLMs in EDA flow automation. 

\minisection{Complex Flow Calls}
This task requires a higher proficiency in EDA tool usage, including traversing parameters, which examines logical reasoning and understanding of each argument of EDA APIs.

\minisection{Parameter Tuner Calls}
This task requires agent systems to provide a parameter-tuning solution, which is a vital step for EDA considering the complexity of the entire process.

\begin{table}[!t]
\centering
\resizebox{1\linewidth}{!}{
\scriptsize
\begin{tabular}{c|c|c|c}
\toprule
\multirow{2}{*}{System} & \multirow{2}{*}{Powered LLM} &  ChatEDA-bench & iEDA-bench \\
& & Acc. & Acc. \\ 
\midrule
ChatEDA & GPT-3.5$^\diamondsuit$ & 28\% & 30\% \\
ChatEDA & GPT-4$^\diamondsuit$ & 62\% & 70\% \\
ChatEDA & AutoMage-70B$^\diamondsuit$ & 74\% & - \\
ChatEDA & AutoMage2-70B$^\diamondsuit$ & 82\% & - \\
\midrule
EDAid & ChipLlama-8B & 88\% & 84\% \\
EDAid & ChipLlama-70B & \textbf{100\%} & \textbf{100\%} \\
\bottomrule
\end{tabular}
}
\begin{tablenotes}[flushleft] \tiny
\item$^\diamondsuit$ The accuracy values of GPT-3.5, GPT-4 and AutoMage models on the ChatEDA-bench are directly cited from the ChatEDA~\cite{wu2024chateda}.
Moreover, we can only evaluate AutoMage models on the ChatEDA-bench due to the unavailability of closed-source models.
\end{tablenotes} 
\caption{The main results of EDA script generation on ChatEDA-bench~\cite{wu2024chateda} and iEDA-bench.}
\label{table:scripteval}
\end{table}

\subsection{Implementation Details} 
We employ QLoRA~\cite{dettmers2024qlora} for the hybrid instruction tuning of ChipLlama models based on Llama3~\cite{dubey2024llama3} models. 
This involves adopting a constant learning rate schedule with a warm-up ratio of 0.03, utilizing the paged AdamW optimizer~\cite{dettmers20218bit} with a learning rate of $1 \times 10^{-4}$, no weight decay, a batch size of 128, and a sequence length of 4096 tokens. 
The models are fine-tuned for 1 epoch on 16$\times$A100 GPUs with 80G memory each.
After hybrid instruction tuning, we obtain ChipLlama-8B and ChipLlama-70B, two expert LLMs for EDA flow automation.
For clarity and differentiation, the single-agent system refers to the divergent-thoughts agent (role $R_{0}$) in EDAid, while the multi-agent system denotes our EDAid. 

\subsection{Main Evaluation Results}
In this study, we integrate LLMs into our EDAid to conduct a comprehensive evaluation. 
According to~\cite{wu2024chateda}, AutoMage models serve as ``brains'' of the ChatEDA system to execute EDA tasks. 
Concurrently, we incorporate GPT-3.5~\cite{brown2020gpt3} and GPT-4~\cite{openai2023gpt4} into the single-agent system (ChatEDA) through their official APIs.
Importantly, only open-source models are utilized within our multi-agent system, as the decision-making process requires calculating the output logits of LLMs.

As demonstrated in \Cref{table:scripteval}, our multi-agent system, EDAid, powered by ChipLlama-70B, achieves the SOTA performance over all other previous SOTA LLM-powered systems, establishing a significant margin.
To explore the generalization of our EDAid on utilizing EDA tools, we also evaluate it on iEDA-bench, which requires LLMs to use EDA tools from the iEDA platform.
Similarly, EDAid powered by ChipLlama-70B also demonstrates outstanding accuracy.
Our multi-agent collaboration system manifests its potential in automating the EDA flow by interfacing with diverse EDA tools from various vendors.

\begin{figure*}[tb!]
    \centering
    \includegraphics[width=0.975\linewidth]{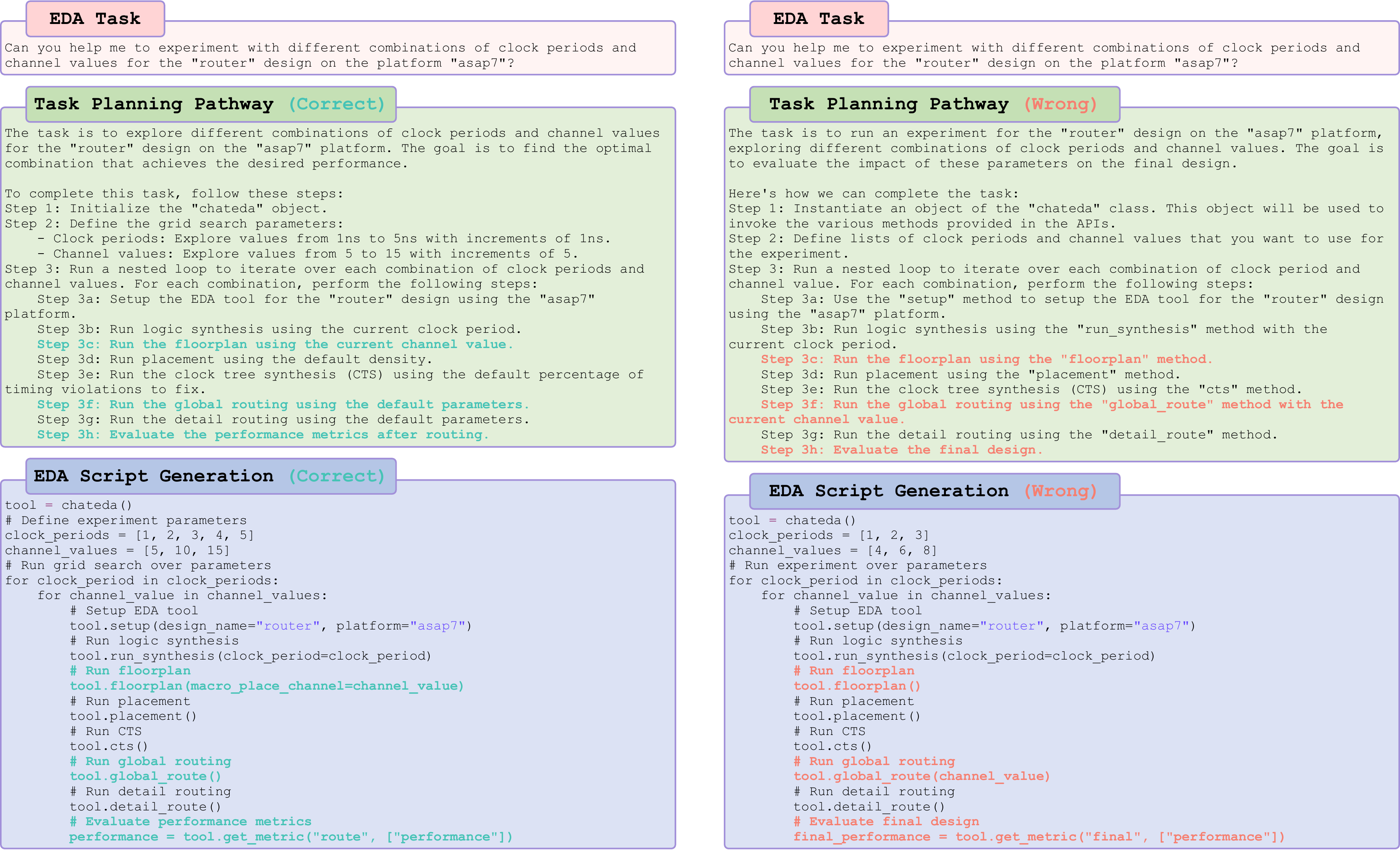} 
    \caption{Divergent thoughts. Right: the divergent-thoughts agent (role $R_{0}$) generates correct task planning pathways and EDA script. Left: the divergent-thoughts agent (role $R_{0}$) makes mistakes in intermediate steps.}
    \label{fig:cases1}
\end{figure*}

\begin{table}[!t]
\centering
\setlength\tabcolsep{2.4pt}
\resizebox{1\linewidth}{!}{
\scriptsize
\begin{tabular}{c|c|c|c|c}
\toprule
\multirow{2}{*}{System} & \multirow{2}{*}{Base LLM} & Hybrid & ChatEDA-bench & iEDA-bench \\
& & Instruction Tuning & Acc. & Acc. \\ 
\midrule
Single-Agent & \multirow{2}{*}{Llama3-8B} & \ding{55} & \textbf{78\%} & 50\% \\
Single-Agent & & \ding{51} & \textbf{78\%} & \textbf{76\%} \\
\midrule
Single-Agent & \multirow{2}{*}{Llama3-70B} & \ding{55} & 88\% & 74\% \\
Single-Agent & & \ding{51} & \textbf{94\%} & \textbf{96\%} \\
\bottomrule
\end{tabular}
}
\caption{Ablation study on hybrid instruction tuning.}
\label{table:ab-hybrid}
\end{table}

\begin{table}[!t]
\centering
\setlength\tabcolsep{2.4pt}
\resizebox{1\linewidth}{!}{
\scriptsize
\begin{tabular}{c|c|c|c|c|c}
\toprule
& &  \multicolumn{2}{c|}{ChatEDA-bench} & \multicolumn{2}{c}{iEDA-bench} \\
\multirow{2}{*}{System} & \multirow{2}{*}{Powered LLM} & \multicolumn{2}{c|}{Acc.} & \multicolumn{2}{c}{Acc.} \\ 
\cmidrule{3-6} 
& & zero-shot & few-shot & zero-shot & few-shot \\ 
\midrule
Single-Agent & GPT-3.5 & 28\% & 56\% & 30\% & 50\% \\
Single-Agent & GPT-4 & 62\% & 82\% & 70\% & 84\% \\
\midrule
Single-Agent & ChipLlama-8B & 74\% & 78\% & 64\% & 76\% \\
Single-Agent & ChipLlama-70B & \textbf{90\%} & \textbf{94\%} & \textbf{90\%} & \textbf{96\%} \\
\bottomrule
\end{tabular}
}
\caption{Ablation study on few-shot prompting and powered LLMs of the single-agent system.}
\label{table:ab-cot}
\end{table}

\begin{table}[!t]
\centering
\resizebox{1\linewidth}{!}{
\scriptsize
\begin{tabular}{c|c|c|c}
\toprule
\multirow{2}{*}{System} & \multirow{2}{*}{Powered LLM} &  ChatEDA-bench & iEDA-bench \\
& & Acc. & Acc. \\ 
\midrule
Single-Agent & \multirow{2}{*}{ChipLlama-8B} & 78\% & 76\% \\
Multi-Agent & & \textbf{88\%} & \textbf{84\%} \\
\midrule
Single-Agent & \multirow{2}{*}{ChipLlama-70B} & 94\% & 96\% \\
Multi-Agent & & \textbf{100\%} & \textbf{100\%} \\
\bottomrule
\end{tabular}
}
\caption{Ablation study on single/multi-agent systems powered by different LLMs.}
\label{table:ab-mas}
\end{table}

\subsection{Ablation Studies}
In the following section, we conduct four ablation studies to further illustrate the effectiveness of our ChipLlama models and our multi-agent collaboration system, EDAid.

\minisection{Hybrid Instruction Tuning}
We compare hybrid instruction tuning and simple EDA-domain instruction tuning and show the results in few-shot scenarios in \Cref{table:ab-hybrid}.
These results underscore the robust generalization capabilities of ChipLlama models after hybrid instruction tuning in utilizing EDA tools across different platforms (e.g. iEDA), notwithstanding its initial training exclusively on EDA tools on the single EDA platform (OpenROAD). 
Moreover, hybrid instruction tuning also achieves improvement on the ChatEDA-bench, which demonstrates that this strategy can also help LLMs to improve their complex reasoning and long-chain tool-calling capabilities.

\minisection{Powered LLMs of Single-agent System}
Considering the powered LLM of each agent in EDAid is vital to reliable EDA flow automation, we test various LLMs by serving them as the controller of the single agent system.
As shown in \Cref{table:ab-cot}, our ChipLlama-70B achieves significant improvements compared to GPT-4 in zero-shot and few-shot scenarios across different EDA platforms.
Notably, ChipLlama-8B also achieves comparable performance to GPT-4, which is impressive considering its model parameters.

\minisection{Few-shot CoT Prompts}
We verify the performance with zero-shot prompts and few-shot prompts and the results are shown in \Cref{table:ab-cot}.
We can observe that GPT models and ChipLlama models demonstrate their capabilities in few-shot learning, which demonstrates the advantages offered by few-shot prompts compared to zero-shot prompts.
Meanwhile, it is worth noting that few-shot CoT prompting is also of great benefit for models' portability across different EDA platforms.

\minisection{Multiple Agents Collaboration}
To assess the efficiency of multi-agent collaboration, we utilize the ChipLlama models to control the agents in our multi-agent collaboration system to perform EDA script generation.
\Cref{table:ab-mas} illustrates that the collaboration of multiple agents brings improvement over the single agent for ChipLlama models. 
This enhancement underscores the capacity of our multi-agent collaboration system to provide dependable assistance in automating the EDA flow.

\section{Case Studies}
In this section, we provide case studies about agents of EDAid, including divergent thoughts agent (role $R_{0}$) and decision-making agent (role $R_{1}$), to demonstrate the core capabilities that enable EDAid to function effectively.

As shown in \Cref{fig:cases1}, the divergent-thoughts agent can generate correct task planning pathways and the corresponding EDA script that can automate the EDA flow successfully.
However, it is common that the divergent-thoughts agent makes mistakes in intermediate steps during the EDA flow automation, which requires long-chain tool-calling capability.
Specifically, the ``global\_route'' method does not have a parameter called ``macro\_place\_channel''.
The ``macro\_place\_channel'' is a parameter for the ``floorplan'' method.
Although most of the task planning pathways are correct, these single errors can still lead to failure in the overall process.

\begin{figure*}[tb!]
    \centering
    \includegraphics[width=0.975\linewidth]{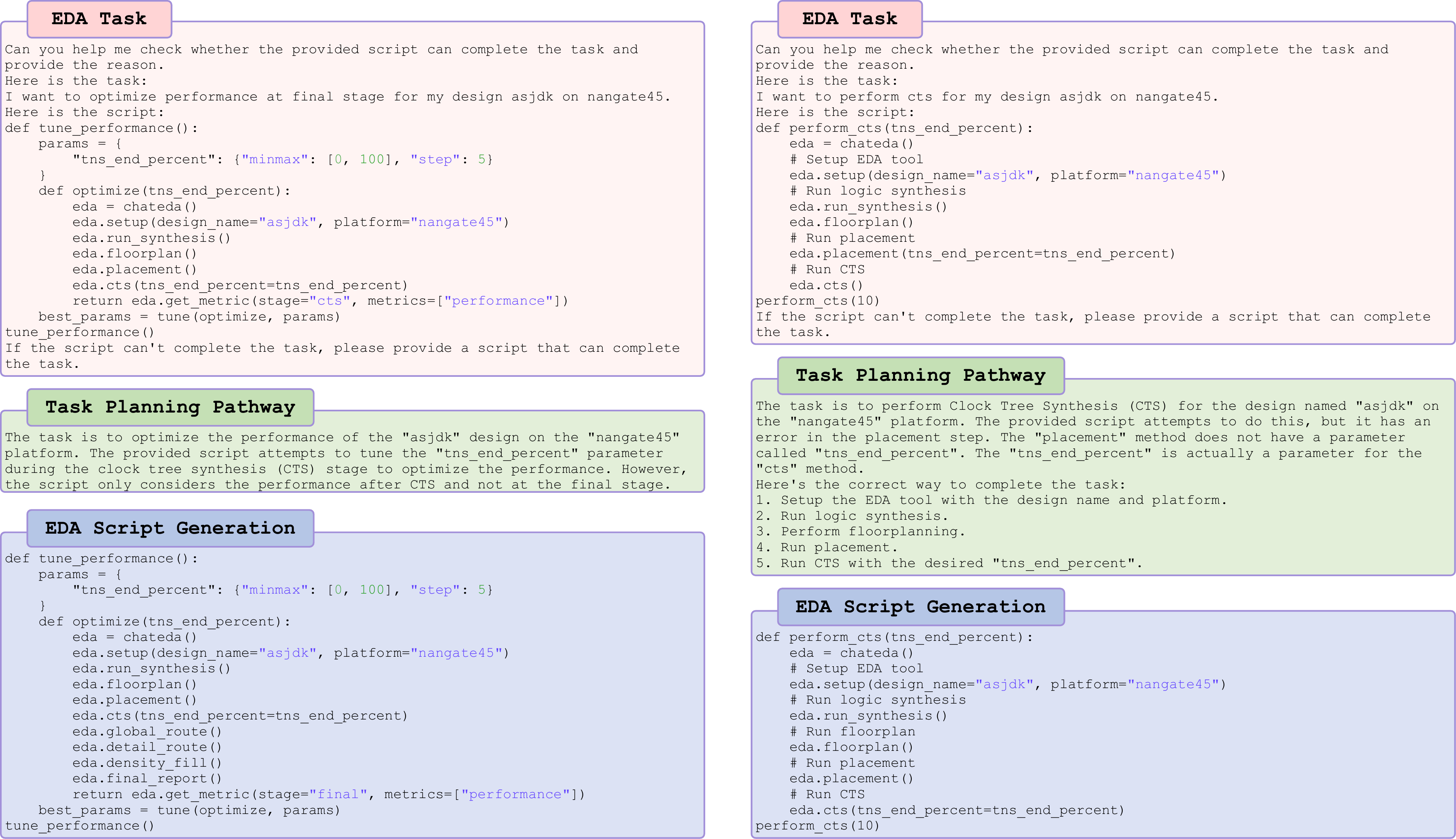} 
    \caption{Error correctness. The decision-making agent (role $R_{1}$) can identify and fix errors in given EDA scripts accurately according to the given EDA tasks.}
    \label{fig:cases2}
\end{figure*}

As illustrated in \Cref{fig:cases2}, the decision-making agent can accurately identify and fix errors of given EDA scripts according to the given EDA tasks.
This capability of error-correct provides a solid base for our decision-making process in our multi-agent collaboration system.
Specifically, the decision-making agent can select the correct EDA script from divergent thoughts (shown in \Cref{fig:cases1}), which guarantees the performance and effectiveness of our multi-agent system, EDAid.

Moreover, we also provide more case studies in \Cref{sec:appendix3}.

\section{Related Works}

\minisection{Chain-of-Thought}
The CoT paradigm~\cite{wei2022cot} encourages LLMs to break down complex problems into several intermediate steps, emulating the way humans reason through a problem. 
Rather than directly outputting the final answer, LLMs are required to generate a step-by-step reasoning process, which can help models handle more complex tasks.
Self-consistency with CoT (CoT-SC)~\cite{wang2022selfcon} improves upon CoT by generating different thought processes for the same problem and the output decision can be more reliable by exploring a richer set of thoughts.
In this paper, we utilize the CoT-SC paradigm to collaborate with various agents in EDAid to perform EDA task planning and script generation for EDA flow automation. 

\minisection{In-Context Learning}
In-context learning \cite{min2022metaicl} has emerged as a transformative paradigm in machine learning, characterized by the meticulous training of models to perform specific tasks through examples and directives provided within an interactive, conversational framework. 
The ability of in-context learning emerges~\cite{wei2022emergent} in large-scale, versatile LLMs~\cite{openai2023gpt4, 2024claude, dubey2024llama3}. 
These models demonstrate an impressive ability to leverage their broad knowledge across various downstream tasks through in-context learning~\cite{brown2020gpt3}.  
In our research, we apply few-shot prompts to enhance the performance and reliability of each agent in EDAid based on in-context learning.

\minisection{LLM-powered Agent System}
Single-agent systems driven by LLMs have demonstrated remarkable cognitive capabilities~\cite{sumers2023cognitive, wang2024agentsurvey, xi2023llmagentsurvey, wu2024chateda}. 
These LLM-powered agents can decompose complex tasks into manageable subtasks~\cite{khot2022decomposed} and methodically think through each component to make better decisions. 
Moreover, the tool-calling capability~\cite{qin2023toolllm} of LLMs enables agent systems to leverage external resources and tools, allowing them to operate more effectively in various scenarios.
Developed based on the single-agent system, several studies~\cite{hong2024metagpt, wang2023unleashing, du2023improving, hao2023chatllm} have enhanced the problem-solving abilities of LLMs by integrating discussions among multiple agents.
The collaboration of multiple autonomous agents, each equipped with unique strategies, can address more dynamic and complex tasks. 
In this work, multiple agents in EDAid collaborate with divergent thoughts to automate EDA flow via complex long-chain EDA tool-calling.

\section{Conclusion}
\label{sec:conclu}
Automating EDA flow by interfacing EDA tools via APIs is imperative to enhance efficiency in electronic design processes.
In this study, we first introduce the ChipLlama-powered agent, collaborating with few-shot prompts for EDA flow automation.
Specifically, ChipLlama models are expert LLMs instruction fine-tuned for the EDA flow automation, which achieve the SOTA performance in EDA script generation and demonstrate versatility across different platforms.
Meanwhile, different few-shot CoT prompts can guide LLMs to generate different task planning pathways for divergent thoughts generation.
Building with divergent thoughts generation and decision-making, we present EDAid, a novel multi-agent collaboration system that utilizes multiple agents for EDA tasks.
This system adeptly handles intricate EDA tasks assigned by designers, thereby automating the EDA workflow effectively.
Our experiments demonstrate the effectiveness of our EDAid.
Moreover, extensive experiments show the significant performance of our ChipLlama models compared to other LLMs in automating the EDA flow.
In conclusion, we anticipate that our work can catalyze the evolution of next-generation EDA tools, inspiring advancements in the field.
\section*{Limitation}
Our multi-agent collaboration system, EDAid, introduces inference latency compared to the single-agent system, considering that the divergent thoughts and decision-making process require multiple inference steps of LLMs.
We will provide more discussion in \Cref{sec:appendix2}.

\section*{Acknowledgement}

This work is partially supported by
The Research Grants Council of Hong Kong SAR (No.~RFS2425-4S02, No.~CUHK14211824, No.~CUHK14210723),
and AI Chip Center for Emerging Smart Systems (ACCESS), Hong Kong SAR.

\newpage
\bibliography{ref/Top,ref/reference}

\newpage
\appendix
\section{Details of Hybrid Instruction Tuning}

\label{sec:appendix1}
We show the proportion of MathInstruct~\cite{yu2023metamath}, CodeInstruct~\cite{wei2023magicoder}, and EDAInstruct~\cite{wu2024chateda} datasets for hybrid instruction tuning of ChipLlama models in \Cref{table:datasets}.

Firstly, the MathInstruct dataset focuses on CoT reasoning, strengthening the logical reasoning skills required for EDA task planning. 
Then, the  CodeInstruct dataset presents complex coding tasks and solutions, offering coding skills for EDA script generation. 
Finally, we utilize the EDAInstruct dataset which provides the EDA domain-specific knowledge and instructions for EDA flow automation.
The composition of these datasets creates a hybrid corpus meticulously designed to expand the capabilities of ChipLlama models. 

\begin{table}[!htbp]
\centering
\resizebox{0.928\linewidth}{!}{
\begin{tabular}{cccc}
\toprule
& MathInstruct & CodeInstruct & EDAInstruct \\ 
\midrule
Proportion & 80K & 100K & 8K \\
\bottomrule
\end{tabular}}
\caption{The proportion of MathInstruct, CodeInstruct and EDAInstruct datasets for hybrid instruction tuning.}
\label{table:datasets}
\end{table}

\begin{figure*}[!tb]
    \centering
    \includegraphics[width=0.975\linewidth]{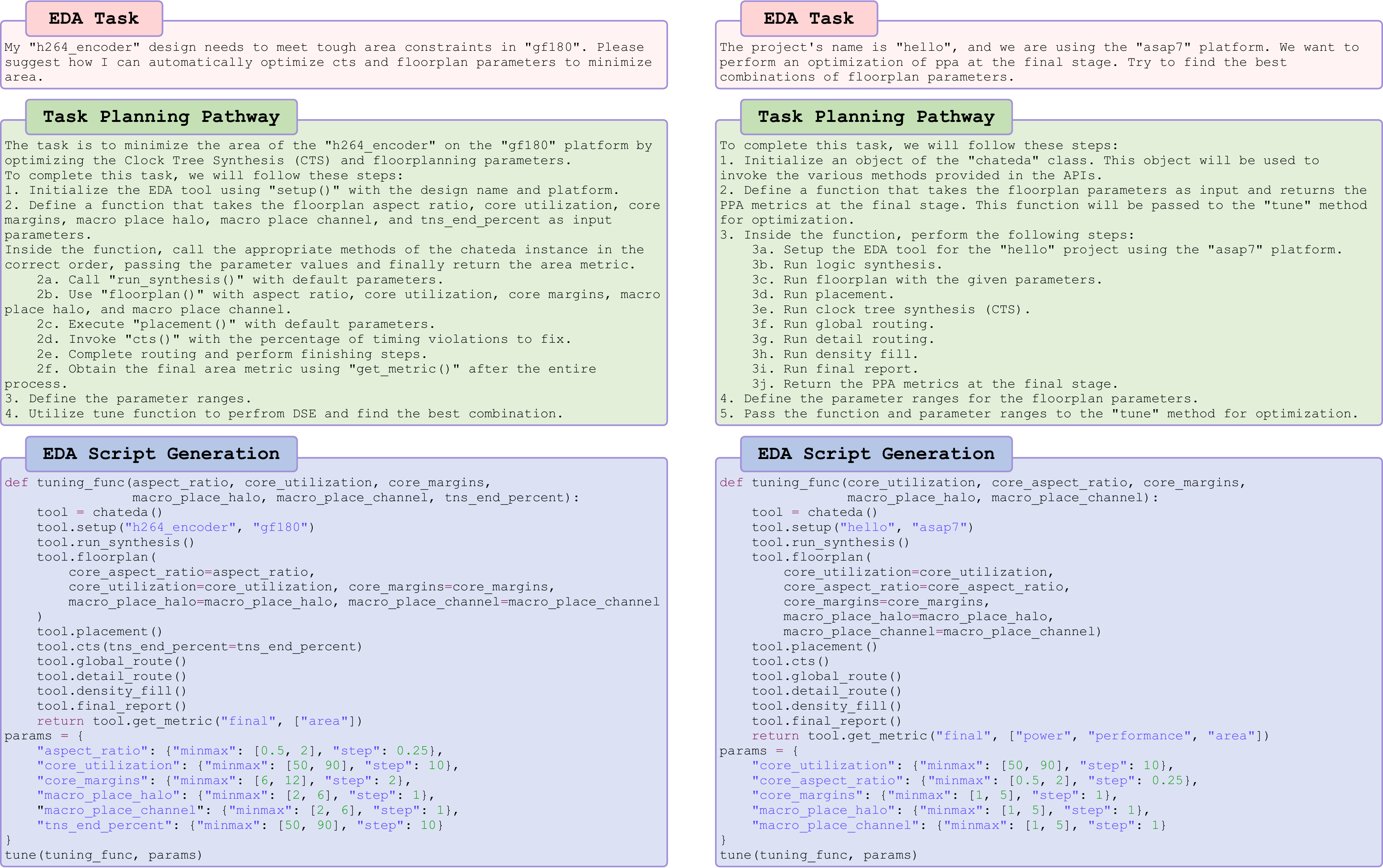} 
    \caption{Case Studies of EDA flow automation with our EDAid powered by ChipLlama models. Each case provides an EDA task, its corresponding task planning pathway and the generated EDA script.}
    \label{fig:cases}
\end{figure*}

\section{Agents in EDAid}
\label{sec:appendix2}
In our experiments, we use three agents to generate divergent thoughts and one agent to compute the probabilities of these thoughts to make the final decision. 
When the number of agents generating divergent thoughts is less than three, the stability and accuracy of EDA flow automation with EDAid improve with the addition of more agents. 
However, once the number of agents reaches three, the performance tends to saturate, meaning that further increases in the number of agents do not necessarily lead to significant improvements in performance. 
In practical applications, considering the costs associated with real-world EDA flows, as we mentioned before, an appropriate increase in the number of agents can still be acceptable if it enhances the system's overall stability and reliability.

\section{More Case Studies}
\label{sec:appendix3}
We provide more case studies to figure out how our EDAid resolves the given EDA task.
As illustrated in \Cref{fig:cases}, we provide two EDA tasks and their corresponding task planning pathways and generated EDA scripts.
Both EDA tasks require the system to provide a parameter-tuning solution.
Our system appropriately grasps the need for the given EDA task and shows an excellent understanding of the details of each API interface parameter.

\end{document}